\documentclass[letterpaper]{article}
\usepackage[preprint]{aaai2027}
\usepackage[hyphens]{url}
\usepackage{graphicx}
\usepackage{xcolor}
\usepackage{natbib}
\usepackage{caption}
\usepackage{algorithm}
\usepackage{algorithmic}

\usepackage{newfloat}
\usepackage{listings}
\DeclareCaptionStyle{ruled}{labelfont=normalfont,labelsep=colon,strut=off}
\floatstyle{ruled}
\newfloat{listing}{tb}{lst}{}
\floatname{listing}{Listing}

\usepackage{booktabs}
\usepackage{colortbl}
\usepackage{amsmath}
\usepackage{amssymb}
\newcommand{\framework}{\textsc{CoCoS}}
\newcommand{\method}{\textsc{CoCoS-GPC}}

\title{Beyond Gene Reconstruction: Learning Cell Representations through Complementary Transcriptomic Views}

\author{
    Jiaqi Xiong\equalcontrib\textsuperscript{\rm 2},
    Yuntao Hu\equalcontrib\textsuperscript{\rm 3},
    Yu Zheng\textsuperscript{\rm 4},
    Yifei Shi\textsuperscript{\rm 5},\\
    Xinyue Guo\textsuperscript{\rm 2},
    Jiaxin Qi\corresponding\textsuperscript{\rm 1}
}
\affiliations{
    \textsuperscript{\rm 1}Computer Network Information Center, Chinese Academy of Sciences, Beijing, China\\
    \textsuperscript{\rm 2}University of Oxford, Oxford, United Kingdom\\
    \textsuperscript{\rm 3}Tongji University, Shanghai, China\\
    \textsuperscript{\rm 4}Hunan University, Changsha, China\\
    \textsuperscript{\rm 5}University of Cambridge, Cambridge, United Kingdom\\
    jxqi@cnic.cn
}

\begin{document}

\maketitle

\begin{abstract}
The rapid growth of single-cell transcriptomic data has enabled the development of foundation models pretrained primarily by reconstructing masked expression values.
This objective encourages these models to learn gene dependencies but does not directly optimize whole-cell representations, which are essential for many downstream tasks.
To bridge this gap, we propose a contrastive pretraining framework that learns cell representations through complementary transcriptomic views.
Since standard contrastive learning is not readily applicable to single-cell pretraining, we introduce specific adaptations along three dimensions --- co-expression-guided gene partitioning, expression-aware contrast-set construction, and competence-gated contrastive onset.
Specifically, we first construct two complementary views of each cell by partitioning its genes according to their co-expression structure. Then, to prevent the model from using gene-set identity as a shortcut, we construct hard negatives by permuting expression values while keeping gene identities unchanged. Finally, we introduce a competence-aware controller to determine how the contrastive objective is applied.
Experiments on cell-type annotation and gene regulatory network inference demonstrate competitive transfer under the evaluated protocols. In the six-network GRN evaluation, our method records the highest mean AUROC and AUPRC point estimates among the compared variants, while the highest-scoring variant differs across individual networks. These results establish complementary-view contrastive learning as an effective direction for single-cell pretraining beyond gene reconstruction.

\end{abstract}

\section{Introduction}
\label{sec:intro}

\begin{figure}[t!]
    \centering
    \includegraphics[width=1\linewidth]{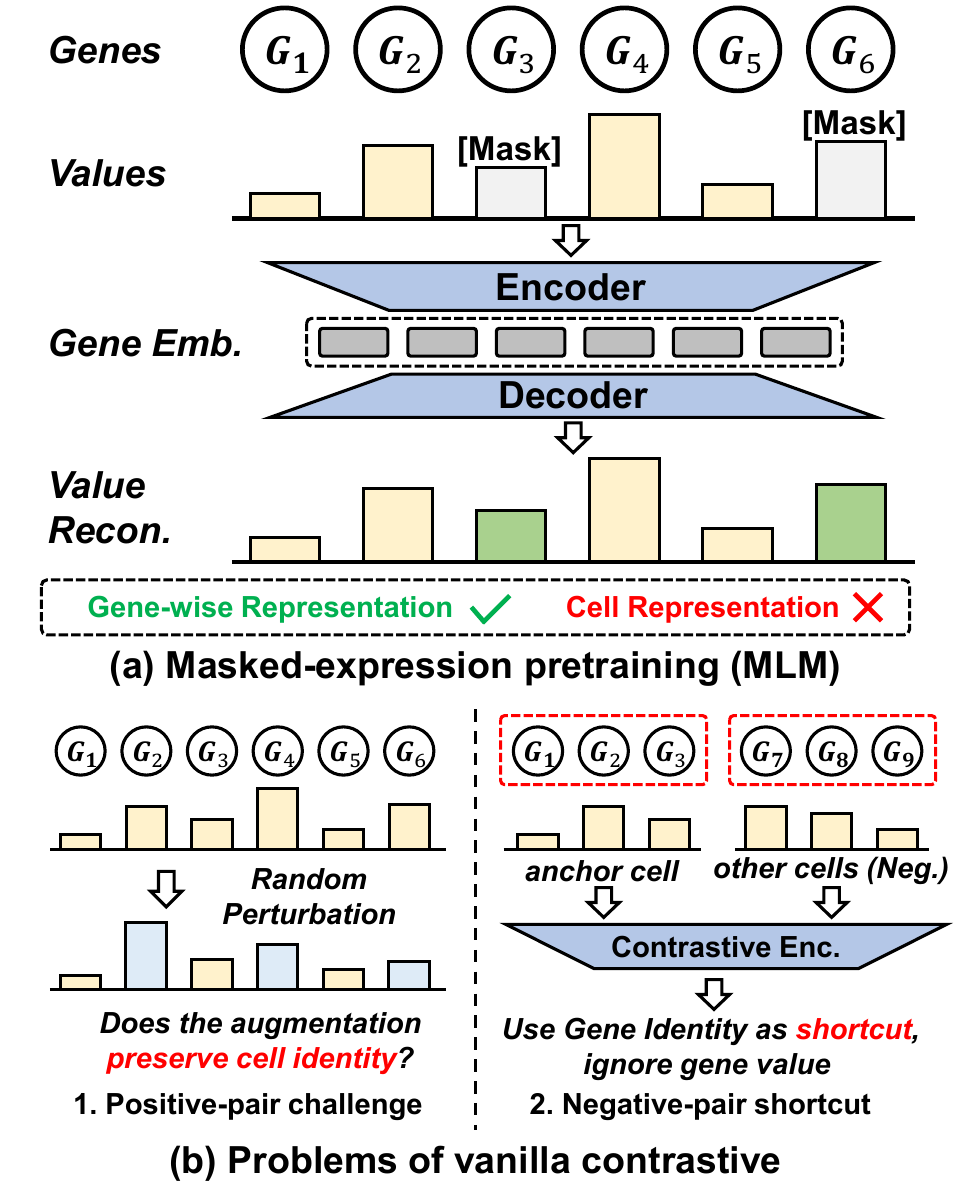}
    \caption{Motivation for \framework{}. (a) Masked expression-value
    prediction directly supervises reconstruction through gene-level features
    but does not impose a metric objective on the whole-cell representation.
    (b) Naively transferring contrastive learning raises two
    transcriptome-specific challenges: unconstrained expression perturbations
    may alter the cellular state used to define a positive pair, while
    negatives with different retained genes may be separated by gene identity
    rather than gene--value correspondence.}
    \label{fig:masked_overview}
\end{figure}

Large-scale single-cell transcriptomic data have motivated the development of foundation models that are predominantly pretrained through masked expression-value prediction~\citep{cui2024scgpt,theodoris2023geneformer,yang2022scbert,hao2024scfoundation}. 
By reconstructing masked values, these models learn gene-level dependencies and capture latent co-expression structure.
However, this objective provides direct supervision only at individual genes, and a masked expression value can often be inferred from a limited neighborhood of correlated genes rather than from a representation of the complete cellular state~\citep{xiong2025crossattention,kotliar2019geneprograms}. As a result, strong reconstruction performance does not necessarily translate into discriminative and transferable whole-cell representations, even though such representations are essential for downstream applications including cell-type annotation and gene regulatory network inference.


This objective mismatch makes cell-level contrastive supervision a natural complement to masked reconstruction.
However, conventional contrastive learning does not transfer straightforwardly to single-cell transcriptomes, where two domain-specific challenges arise, as illustrated in Figure~\ref{fig:masked_overview}.
First, constructing biologically valid positive pairs is nontrivial.
Unlike image augmentations such as cropping, resizing, and color jitter, which can alter low-level appearance while preserving object identity~\citep{chen2020simclr}, perturbing gene-expression values may change cell identity or dynamic biological programs rather than preserve the same biological state~\citep{kotliar2019geneprograms}.
Second, negative-set construction is vulnerable to gene-identity shortcuts.
Conventional contrastive learning typically treats other samples in the minibatch as negatives.
In sparse transcriptomic inputs, however, expression sparsity and sequence truncation may cause different cells to retain different subsets of gene identities.
A model can therefore distinguish negative cells from the anchor using gene-set composition alone, without learning the correspondence between genes and their expression values.


To address these challenges, we introduce \framework{}, a contrastive pretraining framework tailored to single-cell transcriptomes. To construct biologically meaningful positive pairs without altering cell identity, \framework{} keeps measured gene--value pairs unchanged and uses a co-expression-guided gene partition to define two complementary view templates. 
The resulting inputs remain anchored to the same measured cell, avoiding artificial expression shifts while exposing complementary biological evidence. 
To prevent the gene-identity shortcut during contrast-set construction, \framework{} combines expression-aware sampling with same-view negatives and identity-matched hard negatives generated by fixing gene identities and permuting their associated expression values. Because the original and permuted inputs contain the same genes, gene-set composition alone is insufficient, encouraging discrimination based on gene--value correspondence. We refer to these components as \emph{Co-expression-Guided Gene Partitioning} and \emph{Expression-Aware Contrast-Set Construction}, respectively; their motivating challenges are summarized in Figure~\ref{fig:masked_overview}.

While these two components determine what the model contrasts, effective pretraining also depends on when contrastive supervision is introduced. Because the complementary panels share no gene tokens, relating them requires the encoder to recognize that different sets of genes reflect the same underlying biological programs. 
Early in pretraining, however, these cross-gene dependencies have not yet been learned through masked expression reconstruction. Applying the contrastive objective immediately may therefore encourage the model to align the panels using simpler technical cues, such as library size or expression sparsity, rather than their biological correspondence. 
To address this problem, \framework{} introduces \emph{Competence-Gated Contrastive Onset}. The model first learns gene-level dependencies through masked expression modeling. Once label-free diagnostics on held-out cells indicate stable reconstruction competence and cross-view structure, the contrastive objective is introduced gradually. 
Together, the three components address the key challenges of positive-view construction, negative-set design, and contrastive onset, enabling cell-level contrastive pretraining from complementary transcriptomic views.

Experiments across downstream single-cell tasks demonstrate competitive transfer
under the evaluated protocols. Two controlled analyses characterize view
sampling and contrastive scheduling, supporting the effectiveness of
\framework{} for whole-cell representation learning in single-cell
transcriptomic pretraining.

Our main contributions are:
\begin{itemize}

\item We identify the mismatch between masked expression-value reconstruction and whole-cell representation learning, and characterize three requirements for applying cell-level contrast to single-cell pretraining: biologically valid positive views, gene-identity shortcuts in negative samples, and the timing of contrastive supervision.

\item We introduce \framework{}, which addresses these requirements through co-expression-guided gene partitioning, expression-aware contrast-set construction, and competence-gated contrastive onset.
     
    \item We conduct extensive experiments across cell-type annotation and gene regulatory network inference to evaluate the effectiveness of \framework{}. Results demonstrate that properly bounded whole-cell contrastive supervision yields more discriminative and transferable cell representations than pure reconstruction-based methods.
\end{itemize}

\section{Related Work}
\label{sec:related}
\paragraph{Single-cell foundation models.}
Masked reconstruction is common across biological foundation models: DNABERT
and the Nucleotide Transformer predict masked DNA \(k\)-mers, MutBERT models
probabilistic genomic variation, and RNA-FM predicts masked RNA
tokens~\citep{ji2021dnabert,dallatorre2025nucleotide,long2025mutbert,chen2022rnafm}.
Their losses act at corrupted sequence positions, although the learned
representations transfer to sequence-level tasks. Single-cell profiles instead
pair gene identities with expression measurements in sparse cellular
states~\citep{kotliar2019geneprograms}. Large expression corpora have enabled
scBERT, Geneformer, scGPT, and
scFoundation~\citep{yang2022scbert,theodoris2023geneformer,cui2024scgpt,hao2024scfoundation},
with recent benchmark work examining their transfer
behavior~\citep{qi2025scbenchmark}. These models use masked gene or expression
prediction and generative modeling to learn gene dependencies, but do not
explicitly organize whole-cell embedding geometry.

\paragraph{Single-cell representation learning.}
In addition to foundation models, dedicated representation-learning methods
optimize cell embeddings for clustering, integration, and atlas mapping.
Contrastive clustering, CLEAR, and Concerto learn from augmented profiles or
paired encoder views without incorporating contrast into a masked-expression
foundation-model objective~\citep{ciortan2021contrastive,han2022clear,yang2022concerto}.
More recent large-scale approaches bring cell-level contrast into pretraining
in different forms. LangCell combines masked-gene modeling with intra- and
inter-modal contrastive objectives; TABULA jointly performs column-wise gene
reconstruction and row-wise cell contrastive learning; and scConcept replaces
reconstruction with contrastive pretraining over disjoint gene
panels~\citep{langcell,ding2025tabula,scconcept}. Cell-level contrast is
therefore not itself new. \framework{} instead retains masked expression-value
prediction and jointly designs co-expression-guided views, expression-aware
contrast sets with fixed-identity value derangements, and a data-dependent
contrastive onset for this setting.

\paragraph{Adaptive training.}
Curricula order or filter examples by difficulty, with competence-based and
adaptive variants changing exposure according to training progress or model
state~\citep{bengio2009curriculum,platanios2019competence,kong2021adaptivecurriculum}.
GradNorm and Auto-\(\lambda\) instead continuously reweight active
losses~\citep{chen2018gradnorm,liu2022autolambda}. \framework{} controls a
different variable: a fixed, held-out, label-free cohort triggers a previously
inactive contrastive objective while masked-value reconstruction remains
active. This is data-dependent objective onset rather than sample curriculum
or generic multitask loss balancing.

\begin{figure*}[!ht]
\centering
\includegraphics[width=0.98\textwidth]{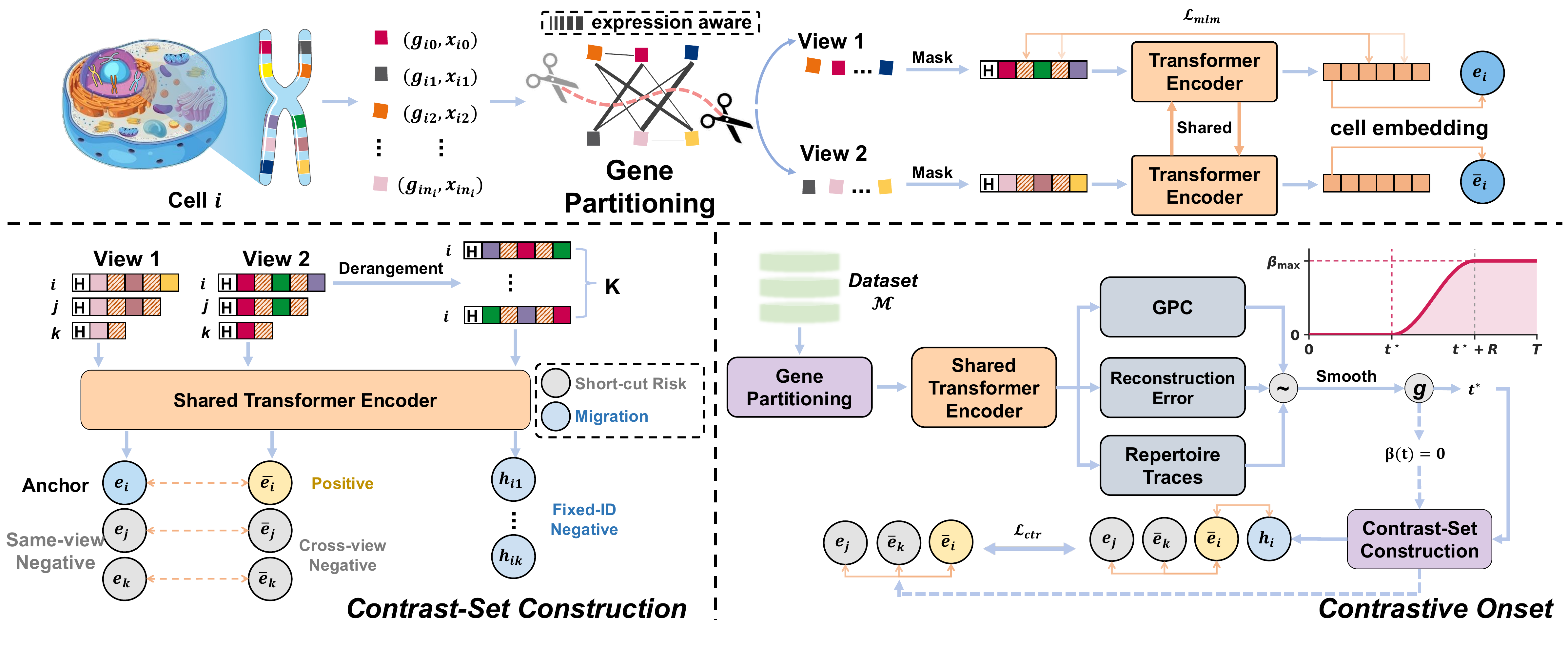}
\caption{Overview of \method{}. Co-expression-guided partitioning and
expression-aware sampling form two disjoint views, which are masked and
reconstructed by a shared Transformer. The paired same-cell views define the
positive, while same-view and cross-view in-batch examples together with
fixed-identity value derangements form the contrast set. A label-free sentinel
tracks smoothed gene-program concordance, reconstruction error, and
representation repertoire; once readiness persists, the controller activates the
contrastive objective and ramps its weight, which remains zero beforehand.}
\label{fig:methodology}
\end{figure*}

\section{Method}
\label{sec:method}

Masked-value reconstruction directly supervises individual gene values but not the geometry of the whole-cell representation. \framework{} augments masked-value reconstruction with three coordinated operations. Co-expression-guided gene partitioning forms complementary views; expression-aware contrast-set construction combines weighted sampling with identity-matched hard negatives; and competence-gated contrastive onset activates cell-level alignment only after the masked predictor satisfies a label-free readiness criterion. The resulting instantiation is \method{}. Figure~\ref{fig:methodology} summarizes the framework, and the following sections introduce the backbone and these operations in training order.




\subsection{Backbone and Masked Pretraining}
\label{sec:backbone}

Let
$\mathcal{M}=\{(\mathbf{g}_i,\mathbf{x}_i)\}_{i=1}^{N}$
denote the pretraining dataset of $N$ cells. For cell $i$,
$\mathbf{g}_i=(g_{i1},g_{i2},\ldots,g_{in_i})$ is its gene-identity
sequence, where $g_{ij}\in\mathcal{G}$ and $\mathcal{G}$ denotes the gene
vocabulary. The aligned expression sequence
$\mathbf{x}_i=(x_{i1},x_{i2},\ldots,x_{in_i})$ contains the corresponding
continuous expression values.

For masked-value prediction, we sample a set of masked positions
$\Omega_i\subseteq\{1,\ldots,n_i\}$ and replace the expression values at
these positions to obtain the masked sequence $\tilde{\mathbf{x}}_i$. We use
the Transformer backbone from scBenchmark~\citep{qi2025scbenchmark} to
encode the gene identities and masked expression values. The masked-value
reconstruction loss in a minibatch $B$ is
\begin{equation}
\mathcal{L}_{\mathrm{mlm}}
=
\frac{1}{\sum_{i=1}^{B}|\Omega_i|}
\sum_{i=1}^{B}
\sum_{j\in\Omega_i}
\left(
f_\theta(\mathbf{g}_i,\tilde{\mathbf{x}}_i)_j-x_{ij}
\right)^2,
\end{equation}
where $f_\theta$ denotes the Transformer encoder together with its
masked-value prediction layer, and
$f_\theta(\mathbf{g}_i,\tilde{\mathbf{x}}_i)_j$ is the predicted expression
value at position $j$. However, because this reconstruction objective provides no explicit supervision for the whole-cell representation, it does not directly optimize the cell embedding for discriminative and transferable downstream use.

\subsection{Our Method: \framework{}}
\label{sec:c1}

\smallskip
\noindent\textbf{Co-expression-Guided Gene Partitioning}.
Constructing biologically valid positive pairs is nontrivial in single-cell
pretraining, because perturbing expression values may alter the underlying
cellular state. \framework{} instead partitions each cell into two complementary
gene views and applies the same masking scheme to both, providing a positive
pair while preserving the reconstruction objective.

We construct an undirected weighted co-occurrence graph
$(\mathcal{G},E,w)$, where $w_{gh}$ counts how often genes $g$
and $h$ co-occur among the most highly expressed genes of a training cell.
After retaining each gene's strongest neighbors, we seek a bipartition that
places strongly co-occurring genes in different views:
\begin{equation}
\max_{\chi:\mathcal{G}\to\{0,1\}}
\sum_{\{g,h\}\in E}
w_{gh}\,\mathbf{1}\{\chi(g)\neq\chi(h)\}.
\end{equation}
We construct the partition with weighted-degree ordering and count-based
placement. Genes are visited in decreasing order of their summed incident
co-occurrence weights; each gene is then assigned opposite the panel containing
more of its already assigned neighbors, with every neighbor contributing one
count irrespective of $w_{gh}$. For each minibatch, we further reverse each
gene's assignment with a small probability and share the resulting partition
across all cells. For cell $i$, the partition defines two complementary sets
of gene positions, $\Omega_i$ and $\bar{\Omega}_i$. The first view masks the
expression values at positions in $\Omega_i$, whereas the second masks those
at positions in $\bar{\Omega}_i$. Therefore, each gene value is observed in
one view and masked in the other, allowing the two views to provide
complementary information while preserving the original gene identities and
expression values. The masked-value reconstruction loss defined above can be
applied to both views.

\smallskip
\noindent\textbf{Expression-Aware Contrast-Set Construction.}
Although complementary masking provides positive pairs, standard in-batch
contrast may exploit gene identities as a shortcut. Different cells in the
same minibatch generally contain different subsets of genes, whereas the two
views of the same cell preserve the same gene identities. The model may
therefore distinguish positive and negative pairs from gene composition alone,
without learning the correspondence between genes and their expression values.

To prevent this shortcut, we combine expression-aware sampling with identity-matched hard negatives.
Specifically, let ${\Omega}_i$ denote the positions kept unmasked in one
view, and let
${\mathbf{x}}_i=(x_{ij})_{j\in{\Omega}_i}$
denote the corresponding visible expression values.
We construct an identity-matched hard negative by permuting the values in
${\mathbf{x}}_i$ while keeping the gene identities $\mathbf{g}_i$ and
the masking pattern unchanged.
The original and permuted views therefore
contain identical gene identities but different gene--value assignments,
requiring the model to discriminate between them based on gene--value
correspondence. The same construction is applied to the complementary view.

Let $\mathbf{e}_i$ denote the
embedding of an anchor view, $\bar{\mathbf{e}}_i$ the embedding of its
complementary view, and $\mathcal{E}_i^{-}$ the corresponding negative set.
The negative set contains embeddings of other cells in the minibatch from
both views, together with the identity-matched hard-negative embeddings
constructed above. For either view chosen as the anchor, the contrastive loss
is
\begin{equation}
\mathcal{L}_{\mathrm{ctr}}
=
-\frac{1}{B}\sum_{i=1}^{B}
\log
\frac{
\exp\!\left(\mathbf{e}_i^{\top}\bar{\mathbf{e}}_i\right)
}{
\exp\!\left(\mathbf{e}_i^{\top}\bar{\mathbf{e}}_i\right)
+
\sum_{\mathbf{e}^{-}\in\mathcal{E}_i^{-}}
\exp\!\left(\mathbf{e}_i^{\top}\mathbf{e}^{-}\right)
}.
\end{equation}
The final objective averages this loss over the two complementary choices of anchor.

\begin{table*}[t]
\centering
{\footnotesize
\setlength{\tabcolsep}{3pt}
\begin{tabular}{@{}l*{11}{c}@{}}
\toprule
Model & MS & Mye & Panc & Lup & Den & IrC & Lep & Mya & Sca & scF & Avg \\
\midrule
scBenchmark~\citeyearpar{qi2025scbenchmark} & 73.51 & 60.98 & \textbf{94.65} & 61.39 & 76.77 & 49.26 & 43.30 & 73.89 & 79.82 & 57.40 & 67.10 \\
scGPT \citep{cui2024scgpt} & 76.48 & \textbf{67.35} & \underline{92.42} & 66.18 & \underline{81.18} & 48.94 & 45.21 & \underline{78.45} & \underline{89.91} & \underline{63.82} & \underline{70.99} \\
Geneformer V2-104M$^{\dagger}$~\citep{theodoris2023geneformer}
  & 71.60 & 64.29 & 89.97 & 71.75 & 79.42 & 51.00
  & 48.86 & 78.89 & 87.39 & 60.64 & 70.38 \\
\midrule
\rowcolor{black!5}
Split-view reconstruction only & \underline{79.92} & 62.77 & 91.72 & 62.37 & 78.85 & 48.81 & \underline{51.97} & 76.00 & 84.89 & 59.73 & 69.70 \\
\rowcolor{black!5}
Contrastive only & 76.15 & 58.90 & 82.50 & \underline{67.13} & \textbf{83.08} & \underline{52.47} & 51.47 & 77.91 & 86.76 & 58.67 & 69.50 \\
\midrule
\rowcolor{black!12}
\textbf{\method{} (proposed)} & \textbf{82.19} & \underline{66.34} & 92.29 & \textbf{72.47} & \textbf{83.08} & \textbf{52.79} & \textbf{60.29} & \textbf{81.03} & \textbf{90.73} & \textbf{66.95} & \textbf{74.82} \\
\bottomrule
\end{tabular}
}
\caption{Cell-type annotation under the unified ten-dataset cell-forward
evaluation protocol: $k$-NN accuracy (\%) from frozen whole-cell embeddings,
one training seed and one downstream split. Unshaded rows are external
references, light-gray rows are internal controls, and the darker row is
proposed. Within each column, all entries attaining the highest value are bold
and the next distinct value is underlined. $^{\dagger}$Geneformer is evaluated
on the same cell rows and seed-42 split after a model-specific top-512
median/rank adaptation. Because this is neither native-input Geneformer nor
numerically interchangeable with the legacy rows, it is shown for coverage but
excluded from the bold/underline ranking.}
\label{tab:main}
\end{table*}

\smallskip
\noindent\textbf{Competence-Gated Contrastive Onset}.
Introducing cell-level contrast before masked reconstruction has learned stable
gene dependencies may encourage the model to align views through technical
cues, such as library size or sparsity. We therefore keep the contrastive
weight $\beta(t)=0$ until a fixed, unlabeled sentinel cohort disjoint from
training indicates readiness. The gate uses neither cell labels nor downstream
scores and is not tied to a predefined epoch.

At probe $p$, let $\mu_p^+$ denote the mean cosine similarity between paired
sentinel views, and let $\mu_p^-$ and $\sigma_p^-$ denote the mean and
standard deviation over mismatched cross-view pairs. We define gene-program
concordance (GPC) and the repertoire score as
\begin{equation}
c_p=
\frac{\mu_p^+-\mu_p^-}{\max(\sigma_p^-,\varepsilon_c)},
\qquad
q_p=
\|\operatorname{std}(\mathbf{Z}_p)\|_2
\sqrt{\operatorname{erank}_p},
\end{equation}
where $\varepsilon_c>0$ is a numerical stabilizer, $\mathbf{Z}_p$ stacks the
normalized sentinel embeddings, and $\operatorname{erank}_p$ is the entropy
effective rank of their centered covariance. GPC measures the separation
between paired and mismatched views relative to off-pair variation, while
$q_p$ monitors embedding diversity.

The same probe also records the pooled masked-value MSE across both views.
Using early sentinel probes for calibration, we smooth the GPC,
reconstruction-error, and repertoire traces. The controller declares
readiness when cross-view concordance is stable and embedding diversity
is preserved for a prescribed number of consecutive probes. The first
optimizer update satisfying these conditions is denoted by $t^\star$, after
which the controller triggers irreversibly.

After the trigger, let $R$ denote the ramp duration and define
$u(t)=\operatorname{clip}((t-t^\star)/R,0,1)$. The contrastive weight is
\begin{equation}
\beta(t)
=
\beta_{\max}
\sin^2\!\left(\frac{\pi}{2}u(t)\right).
\end{equation}
Thus, $\beta(t)$ remains zero before $t^\star$ and increases smoothly to
$\beta_{\max}$ afterward. Identity-matched hard negatives are constructed only when
$\beta(t)>0$.

\smallskip
\noindent\textbf{Overall Objective}.
The complete training objective combines masked-value reconstruction on both
complementary views with the gated contrastive objective:
\begin{equation}
\label{eq:joint-objective}
\mathcal{L}_{\mathrm{overall}}
=
\mathcal{L}_{\mathrm{mlm}}
+
\beta(t)\mathcal{L}_{\mathrm{ctr}}.
\end{equation}
Before the readiness criterion is met, the model is optimized solely through
masked-value reconstruction. After the trigger, contrastive supervision
gradually shapes the whole-cell embedding while reconstruction remains active
on both views. This design preserves gene-level dependency learning throughout
pretraining and introduces explicit cell-level supervision.

\section{Experiments}
\label{sec:exp}

\subsection{Datasets}
\label{sec:datasets}

\paragraph{Pretraining corpus.}
The pretraining corpus contains $1{,}813{,}780$ source rows and follows the data construction of Qi et al.~\cite{qi2025scbenchmark} from the publicly available CELLxGENE collection. Before training,
we reserve three mutually disjoint held-out cohorts: $2{,}048$ rows for
readiness assessment, $2{,}048$ rows for an independent non-controlling audit,
and $10{,}000$ rows for the partition audit. These cohorts are also disjoint
from the training pool, leaving $1{,}799{,}684$ rows for optimization. The
partition-audit cohort evaluates the fixed gene partition and never affects
the contrastive-onset decision.

\paragraph{Downstream evaluation.}
We evaluate frozen whole-cell embeddings on the ten cell-type annotation
datasets assembled by \citet{qi2025scbenchmark} (Table~\ref{tab:main}).
Geneformer V2-104M~\citep{theodoris2023geneformer} is evaluated on the same ten
cell sets, labels, cell rows, and fixed seed-42 splits; only its model-specific
input encoding differs, as detailed in Section~\ref{sec:setup}.
Separately, we score gene-regulatory edges on six BEELINE Specific 1000-gene
networks (hESC, hHep, mDC, mHSC-E, mHSC-GM, and
mHSC-L)~\citep{pratapa2020beeline}, treating all unlabeled candidate pairs as
negatives. Annotation and GRN labels are used only for downstream evaluation;
neither enters pretraining or the contrastive-onset decision.

\begin{table*}[!t]
\centering
{\footnotesize
\setlength{\tabcolsep}{3pt}
\begin{tabular}{@{}l*{11}{c}@{}}
\toprule
Model & MS & Mye & Panc & Lup & Den & IrC & Lep & Mya & Sca & scF & Avg \\
\midrule
scBenchmark~\citeyearpar{qi2025scbenchmark}
  & 86.97 & \underline{70.41} & \textbf{97.89} & 77.94 & 80.38
  & 53.37 & 66.90 & 80.69 & 92.25 & 67.16 & 77.39 \\
scGPT \citep{cui2024scgpt}
  & 84.96 & \textbf{71.50} & 95.95 & \underline{78.63}
  & \underline{84.06} & 53.87 & 57.73 & \underline{81.23} & 92.84
  & \underline{69.48} & 77.02 \\
Geneformer V2-104M$^{\dagger}$~\citep{theodoris2023geneformer}
  & 64.15 & 51.85 & 82.82 & 73.85 & 74.48
  & 48.33 & 65.52 & 77.12 & 81.90 & 42.19 & 66.22 \\
\midrule
\rowcolor{black!5}
Split-view reconstruction only
  & \textbf{88.85} & 69.98 & 97.26 & 77.30 & 82.43
  & \underline{54.39} & \underline{68.09} & 80.25 & \textbf{93.17}
  & 65.03 & \underline{77.67} \\
\rowcolor{black!5}
Contrastive only
  & 81.01 & 64.14 & 87.52 & 72.46 & 82.77
  & 52.34 & 61.30 & 79.41 & 89.63 & 65.34 & 73.59 \\
\midrule
\rowcolor{black!12}
\textbf{\method{} (proposed)}
  & \underline{88.04} & 69.68 & \underline{97.37} & \textbf{79.11}
  & \textbf{84.60} & \textbf{54.77} & \textbf{69.93} & \textbf{82.35}
  & \underline{92.86} & \textbf{69.83} & \textbf{78.85} \\
\bottomrule
\end{tabular}
}
\caption{Linear-probe test accuracy (\%) on the same ten cell-type annotation
datasets as Table~\ref{tab:main}. Evaluation settings and visual conventions
follow Table~\ref{tab:main}; results use one training seed and one fixed
downstream split. $^{\dagger}$Geneformer uses a model-specific top-512
median/rank adaptation and a frozen classifier evaluated only at epoch 50,
whereas the archived legacy rows use their original evaluator. It is therefore
reported for coverage but excluded from the bold/underline ranking.}
\label{tab:linear-main}
\end{table*}

\subsection{Experimental Setup}
\label{sec:setup}

\paragraph{Optimization and training budget.}
The reported \method{} run trains from scratch for $10$ epochs with batch size
$128$ while discarding incomplete final minibatches, yielding $14{,}060$
updates per epoch and $T=140{,}600$ updates in total. We use AdamW with zero
weight decay, mixed precision, and training seed $42$. The learning rate warms
linearly for $10{,}000$ updates and then remains $2\times10^{-4}$. Training
uses one RTX~4090.

\paragraph{Partitioning, contrast-set construction, and onset.}
The proposed configuration uses co-expression-guided gene partitioning and
expression-aware sampling. Each minibatch processes two disjoint gene-panel
views with the same encoder parameters and sums their masked-value MSE terms in
one optimizer update. Thus, the epoch and update budgets are not doubled,
although the separate view and hard-negative encodings are not FLOP-matched to
the single-view reconstruction reference. For each minibatch, we also sample
one shared perturbation mask that independently reassigns each gene to the
opposite view with probability $5\%$. We use a $40\%$ mask ratio, require at least $50$
observed genes per cell, and cap each view at $512$ genes. The contrast set
contains cross-view and same-view in-batch negatives together with $K=4$
paired-view, fixed-identity value derangements. The label-free controller
activates only after three consecutive probes show smoothed concordance above
its calibrated threshold with a small recent slope, reconstruction error below
its reference level, and repertoire above its reference level. All onset
hyperparameters and run-specific calibration values are reported in the
supplementary material.

\paragraph{Frozen-embedding evaluation.}
Under the unified ten-dataset cell-forward protocol, we extract each frozen
whole-cell embedding with maximum sequence length $512$ and
evaluate it using $k$-nearest neighbors ($k{=}10$) and a learned linear probe
on a fixed $0.7/0.3$ train/test split. 

\paragraph{Comparators.}
We report scGPT~\citep{cui2024scgpt} under the same downstream protocol with an
aligned gene vocabulary as an external reference. We additionally report the
public Geneformer V2-104M checkpoint~\citep{theodoris2023geneformer}. After the
common top-512 biological-gene selection, its frozen encoder receives the
model-specific median/rank encoding and outputs a 768-dimensional cell
embedding. This evaluation matches the fixed cell rows, labels, seed-42 split,
L2 normalization, $k$-NN, and final-epoch Linear@50 probes, but not native
full-transcriptome input, pretraining budget, or capacity. Its Linear@50 value
is also not numerically interchangeable with the archived legacy evaluator, so
the row is excluded from the column rankings. The Transformer implementation
released with scBenchmark~\citep{qi2025scbenchmark} is a full-cell
reconstruction-only reference. A split-view reconstruction-only control and a
contrastive-only control provide additional context. The external references
retain their respective pretraining architectures, batch sizes, and whole-cell
readout configurations.
Uniform sampling is the matched comparison for expression-aware sampling.
Table~\ref{tab:view-design} additionally reports dedicated full-cell
MLM, split-view MLM, balanced-random panel, and no-flip view controls.
Three fixed-onset controls compare immediate contrast with one- and two-epoch
reconstruction warmups within a shared training configuration. The proposed
GPC row reports the selected system configuration.
Historical runs that change multiple factors are reported separately as
contextual evidence.

\subsection{Does the Complete Method Learn Transferable Whole-Cell Embeddings?}
\label{sec:main}
Under the unified ten-dataset cell-forward protocol, \method{} records mean
accuracies of $74.82\%$ with $k$-NN (Table~\ref{tab:main}) and $78.85\%$ with
a linear probe (Table~\ref{tab:linear-main}), compared with $70.99\%$ and
$77.02\%$, respectively, for scGPT. Among the ranked legacy rows, under
$k$-NN, \method{} is strictly best on seven datasets and tied for best on
dengue. Within the same ranked set, under linear evaluation, it has the highest
mean accuracy, ranks first on six datasets, and ranks second on three; Myeloid
is the only dataset on which it is outside the top two.
The margin over scGPT is larger for $k$-NN ($+3.83$ points) than for the
linear probe ($+1.83$ points), indicating that the measured gain is more
pronounced in local neighborhood quality than in linear separability under
these evaluators. All values use training seed 42 and downstream split seed 42.

For additional external-model coverage, the daggered Geneformer V2-104M
row~\citep{theodoris2023geneformer} records $70.38\%$ mean $k$-NN accuracy
(Table~\ref{tab:main}) and $66.22\%$ mean Linear@50 accuracy
(Table~\ref{tab:linear-main}). Under this post-selection adaptation, its mean
$k$-NN score exceeds its mean final-epoch linear score by $7.76$ points. We do
not include the row in best/second-best counts or use it for a leaderboard
claim: its median/rank encoding consumes an adapted top-512 view rather than
native full-transcriptome input, and its fixed Linear@50 evaluator is not
numerically interchangeable with the archived legacy linear results.

\begin{table}[!t]
\centering
{\footnotesize
\setlength{\tabcolsep}{3.5pt}
\begin{tabular}{lcc}
\toprule
Variant & Avg.\ $k$-NN & Avg.\ linear \\
\midrule
Full-cell MLM control & 66.01 & 75.19 \\
Split-view MLM control & 64.09 & 76.39 \\
Balanced-random panel & 74.34 & 78.44 \\
No panel-assignment flip & 63.04 & 75.39 \\
\midrule
Uniform sampling & 73.94 & 78.60 \\
\textbf{Expression-aware (proposed)}
  & \textbf{74.82} & \textbf{78.85} \\
\bottomrule
\end{tabular}
}
\caption{Mean $k$-NN and linear-probe accuracies (\%) for view-design variants
under the ten-dataset cell-forward protocol (one training seed and one fixed
downstream split).}
\label{tab:view-design}
\end{table}

\subsection{How Do View-Design Variants Compare?}
\label{sec:view-design}
Table~\ref{tab:view-design} compares six configurations. The proposed
expression-aware variant records the highest mean point estimates
($74.82\%$ $k$-NN and $78.85\%$ linear), while balanced-random panels remain
close ($74.34\%$ and $78.44\%$). Full-cell and split-view reconstruction are
weaker. The no-flip run is also lower but never activates GPC, so it does not
isolate the effect of panel-assignment flips.

The matched sampling pair changes only the within-view sampler.
Expression-aware sampling improves over uniform sampling by $0.88$ $k$-NN and
$0.25$ linear points. Its onset also shifts from $96{,}000$ to $93{,}000$
steps, so the comparison evaluates the sampler within the gated system rather
than at a fixed onset.

\subsection{Do Fixed-Identity Negatives Mitigate the Gene-Identity Shortcut?}
\label{sec:shortcut-diagnostic}

\noindent\textbf{Diagnostic.}
If ordinary in-batch negatives can be separated using gene-set composition,
the model should be highly confident against other-cell candidates but less
confident when a negative preserves the positive's gene identities and changes
only the gene--value correspondence. We measure this behavior with the
ordinary--identity confidence gap: ordinary cross-view positive confidence
minus confidence against four fixed-identity value derangements. A persistent
positive gap indicates that identity-matched candidates remain harder than
ordinary negatives.

\begin{figure}[t]
\centering
\includegraphics[width=0.9\linewidth]{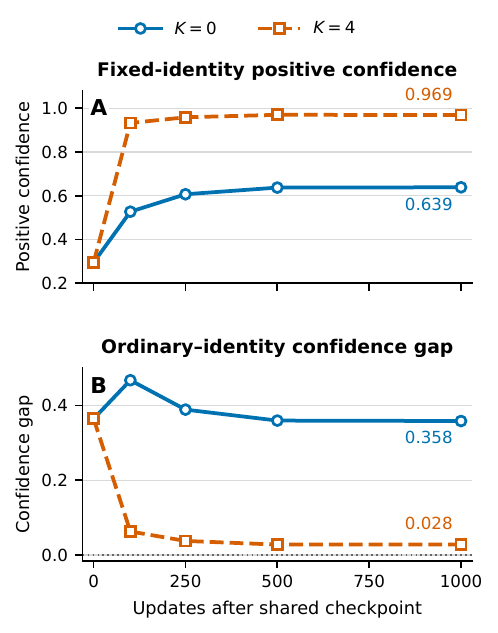}
\caption{Fixed-identity shortcut diagnostic. From a shared checkpoint at step
$84{,}360$, matched $K=0$ and $K=4$ branches run for $1{,}000$ updates and
differ only in fixed-identity negatives. (A) Paired-positive confidence
against four value derangements. (B) Confidence gap between ordinary
cross-view and fixed-identity pools. Both panels use fixed held-out five-way
pools at $\tau=0.07$; this short-fork diagnostic does not estimate downstream
performance or training-seed uncertainty.}
\label{fig:shortcut-diagnostic}
\end{figure}

\noindent\textbf{Answer.}
Starting from the same training state at step $84{,}360$, matched $K=0$ and
$K=4$ branches run for $1{,}000$ updates and differ only in the number of
fixed-identity negatives. At the final probe, $K=0$ is nearly certain against
ordinary negatives ($0.997$ confidence) but is less confident against
fixed-identity derangements ($0.639$), leaving a gap of $0.358$. With $K=4$,
fixed-identity confidence reaches $0.969$ and the gap narrows to $0.028$
(Figure~\ref{fig:shortcut-diagnostic}). Within this controlled diagnostic, the
collapsed gap indicates that fixed-identity negatives mitigate reliance on
gene-set composition and promote sensitivity to gene--value correspondence.
This short fork does not establish downstream gains or training-seed
uncertainty. We also do not compare raw training InfoNCE across $K$, because
$K=4$ adds four denominator logits.

\subsection{When Should Contrastive Learning Begin?}
\label{sec:dyn}

The \method{} controller triggers at step $93{,}000$, after all readiness
conditions hold for three consecutive probes. Before then, the contrastive
weight is zero and neither contrastive gradients nor value-derangement
encodings are computed; afterward, the weight ramps up with the complete
contrast set.

Among the matched fixed-onset controls in Table~\ref{tab:onset-comparison}, a
two-epoch reconstruction warmup performs best, improving over immediate
contrast by $1.50$ $k$-NN and $1.04$ linear points. The GPC row is a contextual
reference rather than part of this matched comparison.

Reevaluating the selected checkpoint over downstream split seeds 42, 43, and
44 gives $74.87\pm0.06\%$ mean $k$-NN and $78.77\pm0.07\%$ mean linear
accuracy.

\begin{table}[!t]
\centering
{\footnotesize
\begin{tabular}{lrrr}
\toprule
Policy & Onset & $k$-NN & Linear \\
\midrule
Immediate contrast & 0k & 72.63 & 77.32 \\
1-epoch warmup & 13.7k & 73.99 & 77.63 \\
2-epoch warmup & 27.3k & 74.13 & 78.36 \\
\midrule
\rowcolor{black!12}
\textbf{GPC (proposed)}
& 93k & \textbf{74.82} & \textbf{78.85} \\
\bottomrule
\end{tabular}
}
\caption{Contrastive-onset comparison under the ten-dataset cell-forward
protocol (training and downstream split seed 42). Onset is in thousands of
optimizer steps. The fixed-onset controls differ only in onset; GPC is the
proposed configuration shown as an unmatched reference.}
\label{tab:onset-comparison}
\end{table}

\subsection{Do the Embeddings Support Gene-Regulatory Edge Prediction?}
\label{sec:biological-transfer}

We evaluate frozen gene embeddings for gene-regulatory edge prediction. Each candidate TF--target pair is represented by the two embeddings, their element-wise product, cosine similarity, and Euclidean distance, and classified using logistic-regression and two-layer MLP probes. Because the task is class-imbalanced, we report both AUROC and AUPRC, with AUPRC more directly reflecting performance on the sparse positive-edge class.

Table~\ref{tab:grn-specific1000-summary} reports results on six Specific
1000-gene networks under the full-candidate negative protocol. \method{}
records the highest mean point estimates among the compared variants for
AUROC (0.8733) and AUPRC (0.7120). Winners vary across networks, so these
averages indicate competitive biological transfer rather than uniform or
statistically significant superiority.

\begin{table}[!t]
\centering
{\footnotesize
\begin{tabular}{lcc}
\toprule
Pretraining variant & AUROC & AUPRC \\
\midrule
scBenchmark~\citeyearpar{qi2025scbenchmark}
    & 0.8715 & 0.7090 \\
scGPT
    & 0.8640 & 0.6962 \\
\midrule
\textbf{\method{} (proposed)}
    & \textbf{0.8733} & \textbf{0.7120} \\
\bottomrule
\end{tabular}
\par}
\caption{Mean AUROC/AUPRC point estimates for GRN edge prediction on six
BEELINE Specific 1000-gene networks with full-candidate negatives, averaging
logistic-regression and MLP probes. Bold marks the highest mean.}
\label{tab:grn-specific1000-summary}
\end{table}

\FloatBarrier

\section{Conclusion}
\label{sec:conclusion}
\framework{} addresses the mismatch between gene-level reconstruction and
whole-cell representation learning by jointly structuring complementary gene
views, contrast sets with identity-matched value negatives, and data-dependent
contrastive onset.
Under the unified ten-dataset cell-forward protocol, the resulting
\method{} embeddings attain the highest mean $k$-NN accuracy among the compared
methods, expression-aware sampling yields higher point estimates than its
matched uniform counterpart, and the held-out controller activates contrastive
learning without downstream labels. The model's frozen gene embeddings yield
the highest mean AUROC and AUPRC point estimates among the compared variants
in the six-network GRN evaluation, although the variant with the highest
point estimate differs across individual networks.
Together, these results support complementary-view contrastive learning as a
useful extension to masked expression reconstruction across cell-type
annotation and gene-regulatory edge prediction, within the evaluated datasets
and protocols.

\section*{Acknowledgments}
This work was supported by the Strategic Priority Research Program of the Chinese Academy of Sciences under Grant No. XDA0460205.

\bibliography{references}

\end{document}